\documentclass{article}

\usepackage{PRIMEarxiv}

\usepackage[utf8]{inputenc} % allow utf-8 input
\usepackage[T1]{fontenc}    % use 8-bit T1 fonts
\usepackage{hyperref}       % hyperlinks
\usepackage{url}            % simple URL typesetting
\usepackage{booktabs}       % professional-quality tables
\usepackage{amsfonts}       % blackboard math symbols
\usepackage{nicefrac}       % compact symbols for 1/2, etc.
\usepackage{microtype}      % microtypography
\usepackage{lipsum}
\usepackage{fancyhdr}       % header
\usepackage{graphicx}       % graphics
\graphicspath{{media/}}     % organize your images and other figures under media/ folder
\usepackage{subfig}
\usepackage{multirow}
\usepackage{tabularx}
\usepackage{amsmath}
\usepackage{natbib}
\setcitestyle{authoryear}
\title{Augmenting Vision-based Human Pose Estimation with Rotation Matrix
}

\author{
  Milad Vazan \\
  Department of Computer and Data Science, Faculty of Mathematical Science\\
 Shahid Beheshti University\\
 Tehran, Iran\\
  \texttt{m\_vazan@sbu.ac.ir} \\
   \And
  Fatemeh Sadat Masoumi \\
  Department of Computer Science, Faculty of Mathematics, Statistics, and Computer Science \\
  University of Allameh Tabataba'i\\
  Tehran, Iran\\
  \texttt{fatemeh\_masoumi@atu.ac.ir} \\
     \And
  Ruizhi Ou\\
    Department of Computer Science at College of Art and Sciences \\
    Boston University\\
    USA\\
    \texttt{ruizhiou@bu.edu}\\
\And
  Reza Rawassizadeh \\
  Department of Computer Science at Metropolitan College \\
  Boston University\\
  USA\\
  \texttt{rezar@bu.edu}
  }

\begin{document}

\maketitle
\large

\begin{abstract}
Fitness applications are commonly used to monitor activities within the gym, but they often fail to automatically track indoor activities inside the gym. This study proposes a model that utilizes pose estimation combined with a novel data augmentation method, i.e., rotation matrix. We aim to enhance the classification accuracy of activity recognition based on pose estimation data. Through our experiments, we experiment with different classification algorithms along with image augmentation approaches. Our findings demonstrate that the SVM with SGD optimization, using data augmentation with the Rotation Matrix, yields the most accurate results, achieving a 96\% accuracy rate in classifying five physical activities. Conversely, without implementing the data augmentation techniques, the baseline accuracy remains at a modest 64\%.
\end{abstract}

% keywords can be removed

\section{INTRODUCTION}
The Covid-19 pandemic has brought sudden changes to individuals and society’s way of life. Lockdown measures have deeply impacted the fitness activities that were regularly practiced in gyms, outdoor spaces (\cite{66}), under expert supervision. Thus, with the suspension and closure of fitness, people were compelled to exercise at home. However, the absence of a trainer and supervision of physical activity hindered this approach. Consequently, conventional training methods for maintaining health and fitness have been replaced by technology (\cite{R1,R2}). Individuals attending the gym can use fitness trackers (either on their wrist or inside their smartphone) or indoor sensors as their training aid (\cite{2}). Nonetheless, if exercises are performed incorrectly, they can be ineffective and even dangerous (\cite{4}). 

Exercise mistakes often arise when users fail to maintain proper form or pose during their workouts. These mistakes can significantly increase the risk of injuries \cite{4}, \cite{25}. To prevent such injuries, receiving guidance and coaching on the correct technique is crucial. Automatically tracking physical activities inside the gym, is an inexpensive technique to facilitate identifying mistakes. Besides, it can reduce the user's cognitive burden while exercising and automatizing the process of fitness data logging. This process can be implemented by an external camera in the user's proximity \cite{R2}.

The use of technology to track and identify human activities is referred to as Human Activity Recognition (HAR). This data is acquired from various devices, such as cameras or wearable sensors \cite{R3}, \cite{x2}. In practice, HAR approaches can be broadly classified into two categories (\cite{x2,x3,x4,x5}): \textbf{vision-based} and \textbf{sensor-based} methods. In the vision-based method, a camera, LiDAR, or similar sensor is placed in the user prxoimity to record the activity from the video stream (\cite{x6}), eliminating the need for users to wear sensors \cite{x1}. 

This research aims to focus on classifying physical activities collected by camera. However, the significant challenge associated with this task is the high requirement for a large and accurately labeled dataset for training. The process of manually collecting and labeling large amounts of data is resource-intensive and time-consuming. To address this challenge, we proposed a novel approach that utilizes only one sample from each activity. 

. Since the camera captures data from a single source and users are positioned toward the camera, the absence of multiple samples creates an ambiguous view for the algorithm. This issue is highlighted when another layer of machine intelligence, such as pose estimation, is added on top of the raw video data. As a result, the HAR algorithm may struggle to detect the correct activity due to limited insight from different angles. To mitigate this issue, we introduce different camera perspectives on physical activities, which enhance the algorithm's efficiency. In particular, to enhance the pose information obtained from the video, we propose a data augmentation method based on the Rotation Matrix (\cite{x8}). This approach helps to alleviate the ambiguity caused by differing camera perspectives.

In this study, we experimented with different pose estimation algorithms and identified BlazePose (\cite{38}) as the preferred model to accurately derive the crucial human body key points from every video. Explaining the details of our experiments for pose estimation algorithm selection is not in the scope of this paper. After we apply our augmentation, we experiment with different machine learning models and report their accuracy. 

The contribution of this research can be summarized as follows: 
\begin{itemize}
    \item We tackle the issue of time and labor-intensive data collecting and labeling by adopting a single sample per class approach. 
    \item We propose a Rotation matrix as a data augmentation method to mitigate the challenges associated with variable camera perspectives. 
    \item By comparing several classification algorithms, We identify SVM with SGD optimization (SVM-SGD) as the most efficient machine learning model for classifying five activities, achieving an impressive accuracy rate of 96\%.
\end{itemize}

\section{Literature Review}
Since our approach uses pose estimation data collected by video from a third person's view, and data augmentation in computer vision, we describe related work in each section separately. In particular, we review \textbf{Vision-based Human Pose Estimation}, \textbf{Vision-based workout analysis}, and \textbf{data augmentation approaches} used on pose data.
\subsection{Vision-based Human Pose Estimation}
Human Pose Estimation (HPE) is one of the applications of computer vision. It refers to a technique for obtaining the spatial coordinates of human joints on a person's body in images or videos (\cite{5,14}). HPE can assist in understanding and retrieving information about human activity (\cite{6}). Human behavior understanding, violence detection, athletics activity analysis, and generating performance feedback are a few of the many areas in which it has immense applications (\cite{7,25}). Based on the dimensionality of the output, there are two types of human pose estimation: \textbf{2D pose estimation} (\cite{d1,d2,d3,d4,d5}) and \textbf{3D pose estimation} (\cite{d6,d7,d8}). 2D pose estimation refers to predicting the key points appearing in the image. In 3D pose estimation, key points are arranged in three dimensions as an output (\cite{27}). A major challenge of 3D pose estimation is the loss of depth information caused by occlusion and blur in the usual intensity images (\cite{28}).

From another perspective, Human pose estimation can be classified further according to the number of individuals in an image into \textbf{single-person human pose estimation} and \textbf{multi-person human pose estimation}. A single-person human pose estimate describes the task of computing the human skeletal key points from an image or video frame (\cite{29}). On the other hand, a multi-person pose estimate includes estimating the key points of multiple individuals, in which the number of individuals is not known beforehand (\cite{30}). To estimate the poses of more than one individual, it is necessary to distinguish the poses of each person (\cite{11}). However, recovering the absolute pose of a 3D object in camera-centered coordinates is challenging. The inherent depth ambiguity and occlusions present in cluttered scenes make it challenging for multiple instances to estimate their precise locations \cite{12}.

The pose estimation process can be divided into two major approaches: \textbf{bottom-up} and \textbf{top-down} approaches (\cite{8,9,16,17,18,19,20}). By fitting human body models to input images, bottom-up methods estimate all parts of the human body (\cite{8}\cite{20}). In other words, these methods detect body parts first and then assemble them into an object that looks like a human (\cite{15}). These methods generally operate in two steps (\cite{10,22}):  first, body joints are detected without knowing how many people are present or where they are located. Next, a person's postures are formed by grouping detected joints. By using the target detection algorithm to locate the human target contour, the top-down method is used to find the human key points in the body contour using the human bone key points detection algorithm (\cite{17}). These methods also operate in two steps (\cite{13,21,22}): first, detecting the person, and in the next step, estimating their pose based on the cropped area.

As a result of the bottom-up approach, all of the key points are detected and grouped before human joints are assigned key points. In contrast, the top-bottom method operates in the opposite manner. It detects the human first and then predicts the person's key points (\cite{23}). Multi-person pose estimation generally exhibits better accuracy with top-down methods, while bottom-up methods are better at controlling inference time (\cite{19}). In summary, the bottom-up pose estimation approaches make it difficult to merge estimations of human body parts. In contrast, the top-down approach has the disadvantage that the calculation cost increases as the number of people increases (\cite{18}).

\subsection{Human Pose Data Augmentation}
It is a common practice for computer vision applications to augment image datasets with synthetic data to gain better accuracy in a machine learning task (\cite{XX1}). Human pose estimation is another area for data augmentation, but the lack of diversity in existing human posture datasets restricts the generalization and, thus, augmentation methods.

PoseAug \cite{x14} is an augmentation framework, which learns to augment available poses towards a greater variation in geometry aspect by using a differentiable procedure. In this approach, the estimation errors are taken as feedback, and more robust poses are generated. This improves the generalization, leading to an increase in estimation performance. This method has been applied to Human3.6M (H36m)  (\cite{n1}), MPI-INF-3DHP (3DHP) (\cite{x17}), 3DPW  (\cite{n3}), MPII  (\cite{n4}), and LSP  \cite{n5} datasets. It improves the model performance with different amounts of training data on H36M and 3DHP, this types of improvement is useful where there is limited training data. 

\cite{xy15} designed a data augmentation approach that conducts pose network estimation and data augmentation simultaneously by applying a reward/penalty on a 3D HPE network to optimize it. Pre-trained information is used to create cross-distribution of variation. This approach gained higher performance in contrary to some state-of-the-art methods with both weakly (\cite{x30,x31,x32,x33,x34}) and fully supervised (\cite{x36,x37,x38,x39}). Authors built an autoencoder that learns the given pose and generates new poses based on the same style. This generated pose is input to a generative adversarial network (GAN). Their method can generate new labeled data from small labeled data, and it was applied to the Human3.6M dataset (\cite{x16}). This 3D pose estimation method increases the neural networks’ accuracy to 56\% on the PCK0.2 dataset, and with regards to 2D pose estimation, it increases the accuracy to 69\% on the PCK0.2 dataset.

\cite{x21} presented a synthesis method using Motion Capture (MoCap) data to augment a real-time 2D pose dataset. Their approach generates synthetic images by mixing annotated images, providing equivalent 3D pose annotations. Initially, the 3D poses are clustered into \emph{K} classes, and then a CNN classifier with K classes is trained to predict pose class distributions given an individual's bounding box. The evaluation was conducted using the protocol introduced in (\cite{x19,x18}), considering six subjects (S1, S5, S6, S7, S8, and S9). The protocol was tested under three conditions: (i) using synthetic data from H3.6M dataset, (ii) using real images of 190,000 poses, and (iii) using both synthetic and real images combined. Ultimately, the classifier achieved a performance of 97.2 (mm) error when trained on either synthetic or real datasets, and an 88.1 (mm) error when trained on both datasets together.

\cite{x22} suggested an evolutionary framework improving the generalization of 2D-to-3D networks for human pose estimation. As Human 3.6M (H36M) dataset (\cite{n1}) was used, and the model’s performance was evaluated by two protocols in Mean Per Joint Position Error (MPJPE) in millimeters. In the first protocol (P1), MPJPE is directly computed. In the second protocol (P2), the ground-truth 3D poses are aligned with the predictions using a rigid transformation before calculating the MPJPE. Another variant of the evaluation protocol, referred to as p1*, incorporates ground truth 2D key points as inputs and eliminates the impact of the initial stage model. Furthermore, MPI-INF-3DHP (\cite{x17}) was applied as a benchmark to evaluate the proposed model's generalization in unseen environments. MPJPE method in comparison with weakly-supervised (\cite{x23,x24,x25,x33}) and fully-supervised approaches\cite{x27,x36,x37,x30,x38,x39},for all actions in H36M (\cite{n1}) gained the lowest error for P1, with 60.8 (mm) and 50.09(mm), and for P1*, with 50.05 (mm) and 34.5 (mm), respectively. In terms of p2, the weakly-supervised method gained a 46.02 (mm) error, while the fully-supervised method gained 38.0 (mm), which was more than the achieved error by Yang et al. method (\cite{x36}), with 37.7 (mm).

\subsection{Vision-based workout analysis}
Artificial Intelligence (AI) and Wearable technologies show a significant impact on sports tracking and coaching  (\cite{R2}). Computer vision and Natural Language Processing (\cite{XR}) play a particularly important role among these technologies. AI-enabled applications can effectively monitor and enhance physical exercises, providing valuable assistance to the end users. In the following, we list some examples of how machine learning benefits physical activity tracking. 

\cite{39} developed a real-time deep learning system capable of recognizing yoga poses performed by a user and providing visual guidance through a smartphone camera. Their system incorporates two modules: a pose estimator that employs OpenPose (\cite{openpose}) to identify 25 key points on the user's body, and a pose detector that uses a deep neural network model with a time-distributed CNN layer and a long short-term memory (LSTM) layer. The outputs of these layers are passed through a SoftMax function and transmitted to a dense layer. The chosen OpenPose model achieved an accuracy of 99.87\% and 99.91\% for the training and test datasets, respectively.

\cite{40} have proposed a hybrid deep learning model for real-time yoga pose recognition using a standard RGB camera. This model includes a time-distributed CNN layer that extracts features from key points in each frame detected by OpenPose, followed by an LSTM layer for temporal prediction. The use of LSTM improves the model's robustness and reduces error by taking the pattern of the last frame into account. The model outputs "No Asena" when the SoftMax value for the majority of the predictions in a series is less than the threshold value. After analyzing 45 frames, the model detects yoga poses with a frame-wise accuracy of 99.04\% and an overall accuracy of 99.38\%. In real-time testing with 12 individuals, the model achieved an accuracy of 98.92\%.

In a study\cite{43}, a physical form correction framework that utilizes CNN to classify images of individuals performing fitness exercises as either correct, e.g., hips too low, or hips too high. Then, their framework provides live feedback to the user to correct their posture. A labeled dataset of 2400 images of two static fitness exercises was used to train their model. The CNN model achieved an impressive 99\% accuracy in real-time.

 \cite{44} developed a real-time indoor error detection system by analyzing the angle between limb pairs and providing corrective actions to the user. They utilized a CNN model on the COCO dataset \cite{lin2014microsoft} to extract the coordinates of body parts. To compare the user's performance with the trainer's action using key-frame mapping, videos of the two individuals were recorded and synchronized using Dynamic Time Wrapping (DTW) \cite{muller2007dynamic}. However, there is no information on the accuracy of this method is provided.

In another study\cite{x11}, researchers presented a camera-based system that utilizes an end-to-end pipeline, to simultaneously detect, recognize, and track multiple individuals performing exercises. Their system can accurately segment exercises with an 84.6\% accuracy rate, recognize five different exercise types with 93.6\% accuracy, and count repetitions within an average error margin of +-1.7. However, this system may face two potential challenges: (1) Two different exercises may be clustered together in a single segment, and (2) One exercise may be clustered across two different segments.

\cite{x12} proposed a system that performs three tasks: person's pose estimation, exercise recognition, and repetition counting.  In their study, the heatmaps matrices were found to be effective in identifying the movement and distribution of the body joints. The ResNet34-based network (\cite{he2015deep}) was used for exercise recognition and achieved 95.69\% accuracy. This approach is designed for single-person detection and can estimate seven human poses.

Table \ref{tab:activity-recognition}  summarizes the previous works. Generally, the earlier studies aimed to tackle the following issues:
\begin{enumerate}
    \item A significant portion of previous research on motion detection has concentrated on yoga movements (\cite{39} \cite{40}), which are relatively simpler to detect than more intricate fitness movements.
    \item The number of their training samples has been large and, therefore, hard to generalize for different types of physical activities.
\end{enumerate}
\begin{table}[htbp]
\centering
\caption{summarizes the previous works}
\label{tab:activity-recognition}
\resizebox{\textwidth}{!}{%
\begin{tabular}{|l|l|l|l|c|c|l|c|}
\hline
\textbf{Reference} & \textbf{Method} & \textbf{Data type} & 
\begin{tabular}[c]{@{}l@{}}\textbf{Type of}\\  \textbf{Activity}\end{tabular}
 & 
\begin{tabular}[c]{@{}c@{}}\textbf{Number of}\\  \textbf{activities}\end{tabular}
 & 
\begin{tabular}[c]{@{}c@{}}\textbf{Number of} \\ \textbf{samples}\end{tabular}
 & \textbf{Keypoints Detection} & \textbf{Accuracy} \\ \hline
\cite{39} & CNN-LSTM & video & Yoga & 6 & 88 videos & OpenPose, MaskRCNN & 99.91\% \\ \hline
\cite{40} & CNN-LSTM & video & Yoga & 6 & 88 videos & OpenPose & 99.38\% \\ \hline
\cite{43} & CNN & image & Fitness & 2 & 2400 images & - & 99\% \\ \hline
 \cite{44} & DTW & video & Fitness & 2 & - & 
\begin{tabular}[c]{@{}l@{}}Realtime Multi-Person\\  2D Pose Estimation using\\  Part Affinity Fields \textbackslash{}\cite{x8}\end{tabular}
 & - \\ \hline
 \cite{x11} & multilayer perceptron & video & Fitness & 5 & 412 videos & - & 93.6\% \\ \hline
 \cite{x12} & ResNet34 & video & Fitness & 7 & 29187 videos & MSPN \cite{x20} & 95.69\% \\ \hline
\end{tabular}%
}
\end{table}

Our research proposes a model by using SVM with a Stochastic Gradient Descent (SGD) optimizer. It can estimate five fitness activities. For keypoint detection, Blasepose \cite{38} is used, and the model achieves an accuracy of 96\% with 30 videos as the sample. Our cost-effective approach minimizes labeling expenses by using a single sample per class and augmenting it with rotation matrix data. This resolves challenges with camera orientations and placement ambiguity. Consequently, the implementation of this methodology led to a substantial enhancement in the classifier's performance, raising its result output from 64\% to 96\%.

In this study, we focus on complex fitness movements using only \emph{one} training sample for each class and employ data augmentation techniques to enhance the accuracy.

\section{Materials and Methods}
\subsection{Overview}
\begin{figure}[htbp]
\includegraphics[width=\textwidth]{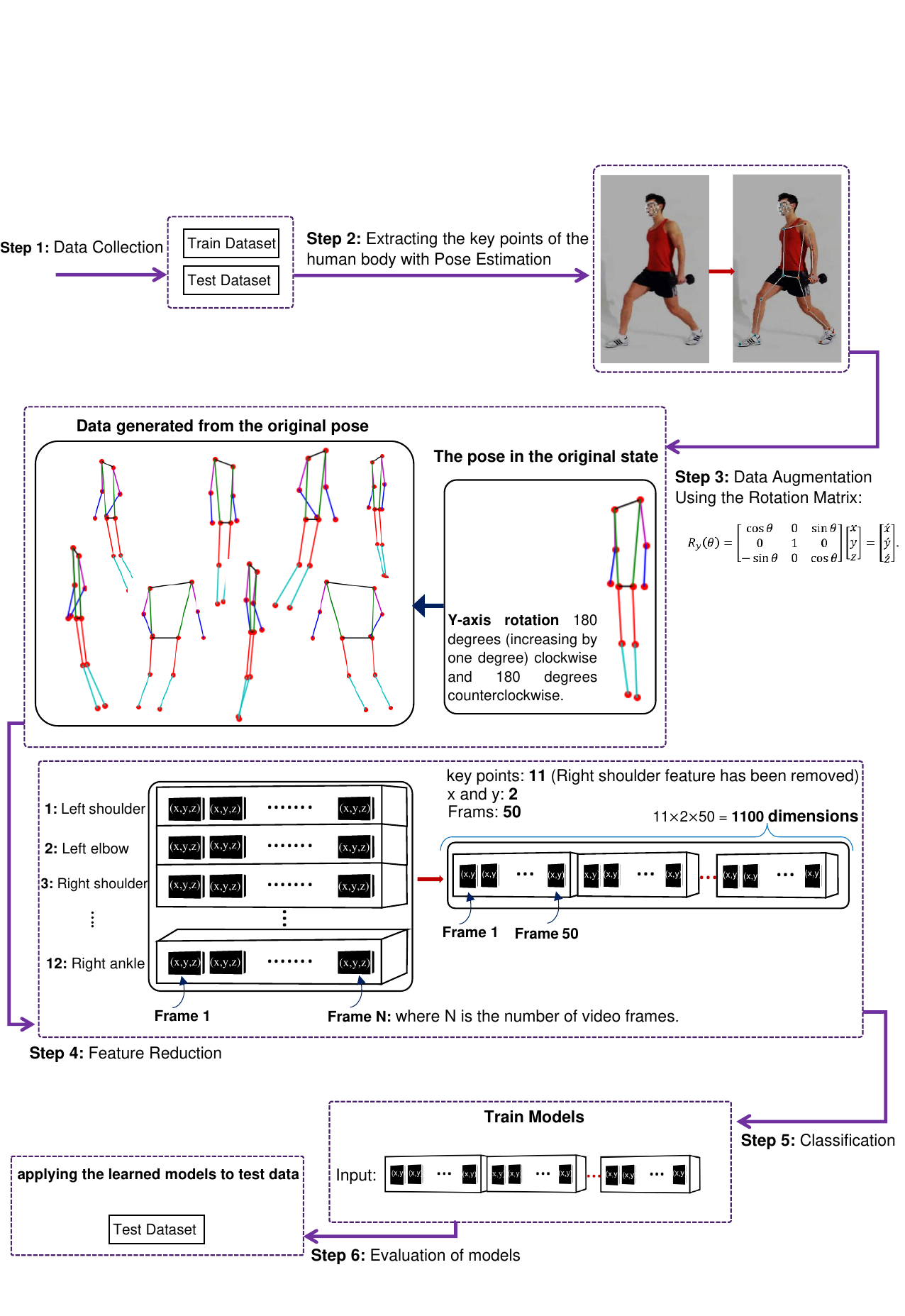}
      \centering
      \caption{Workflow diagram of the proposed method}
      \label{fig: Workflow}
\end{figure}

Our method aims to employ computer Vision and classify physical activities. Figure \ref{fig: Workflow} depicts the method's workflow.  After receiving each video and extracting the body's key points through pose estimation, our data augmentation is performed on each video using a rotation matrix.  Afterward, the data undergoes a feature reduction process before model training. This step aims to decrease the data's complexity to a more manageable size, enhancing efficiency in the training process while preserving accuracy. A detailed introduction to each component will be presented in the subsequent sections.
\subsection{Exercise Types and Data Collection}
We considered five types of exercise: angled leg presses, Chin-ups,  dumbbell lunges, hack squats, and squats. Our rationale for selecting these experts is to have a sample exercise for each body part. Then, to train the dataset, we collected one sample of each class from Youtube, Instagram or TikTok and checked with a reference video that is evaluated by the human. This ground truth video is also collected by human. For the test dataset, we manually collected and labeled videos for each class in the number that can be seen in Table \ref{tab:data}.
\begin{table}[htbp]
\centering
\caption{Details of the test dataset}
\label{tab:data}
\begin{tabular}{llllll}
\hline
\multicolumn{1}{|c|}{\textbf{Class}}             & \multicolumn{1}{c|}{angled leg presses} & \multicolumn{1}{c|}{chin-ups} & \multicolumn{1}{c|}{dumbbell lunges} & \multicolumn{1}{c|}{hack squats} & \multicolumn{1}{c|}{squats} \\ \hline
\multicolumn{1}{|c|}{\textbf{number of samples}} & \multicolumn{1}{c|}{6}                  & \multicolumn{1}{c|}{5}        & \multicolumn{1}{c|}{6}               & \multicolumn{1}{c|}{4}           & \multicolumn{1}{c|}{4}      \\ \hline 
\end{tabular}
\end{table}
\subsection{Pose Estimator and Create a dataset}
We used BlazePose (\cite{38}), a pre-trained model for extracting the key points of the human body. BlazePose is a lightweight convolutional neural network architecture for mobile devices and can be sped up to super-real time on a mobile Graphic Processing Unit (GPU). The network identifies 33 key points of body parts of an individual. This body pose estimation neural network uses heatmaps and regression to coordinate key points. The design of this pipeline employs a detector-tracker setup, and it consists of a lightweight body pose detector followed by a pose tracker network. The tracker predicts the keypoint positions; if the tracker indicates that there is no person in the scene, the detector will be re-run on the next frame. 
\par
\begin{figure}[ht!]
\includegraphics[scale=0.6]{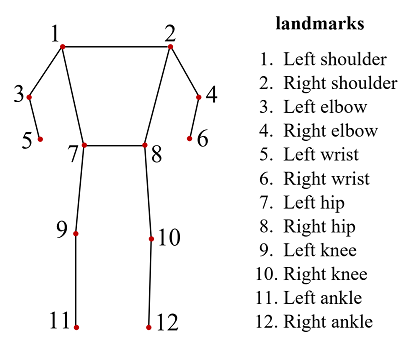}
      \centering
      \caption{BlazePose landmarks}
      \label{fig: bz}
\end{figure}

As mentioned earlier, the coordinates (x,y,z) of 33 skeleton key points can be calculated using this model. However, for this study, we did not need all these key points and used only 12. Selected key points include: (1) left\_shoulder, (2) right\_shoulder, (3) left\_elbow, (4) right\_elbow, (5) left\_wrist, (6) right\_wrist, (7) left\_hip, (8) right\_hip, (9) left\_knee, (10) right\_knee, (11) left\_ankle, and (12) right\_ankle (See Figure \ref{fig: bz}).
\par
A Blazepose model is available in three variations: lite, full, and heavy. From lite to heavy, model accuracy increases, and inference speed decreases. We chose the heavy model because accuracy is important for our problem.

\begin{figure}[h]
    \centering
    \subfloat[\centering X]{{\includegraphics[width=7cm]{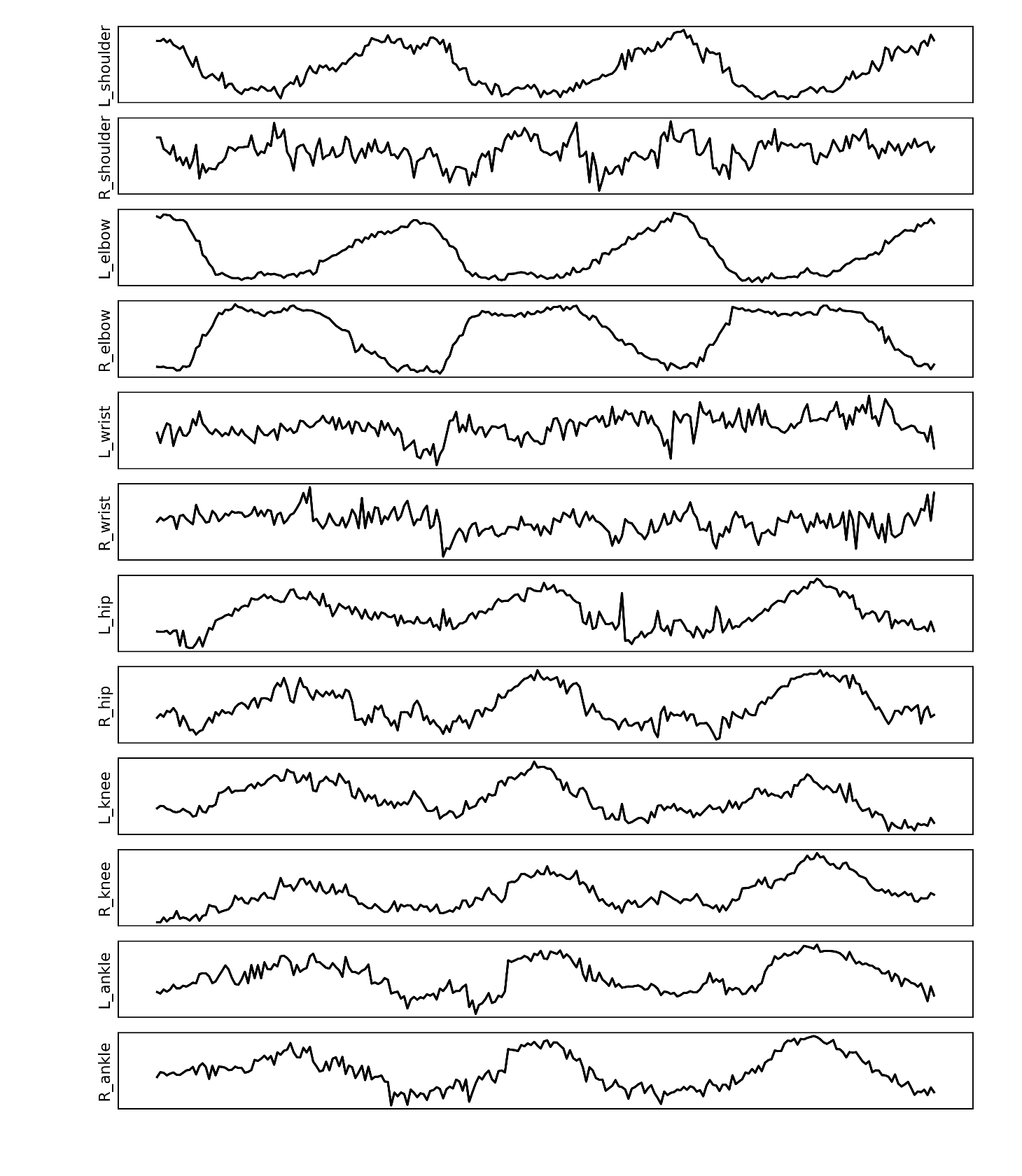} }}%
    \qquad
    \subfloat[\centering Y]{{\includegraphics[width=7cm]{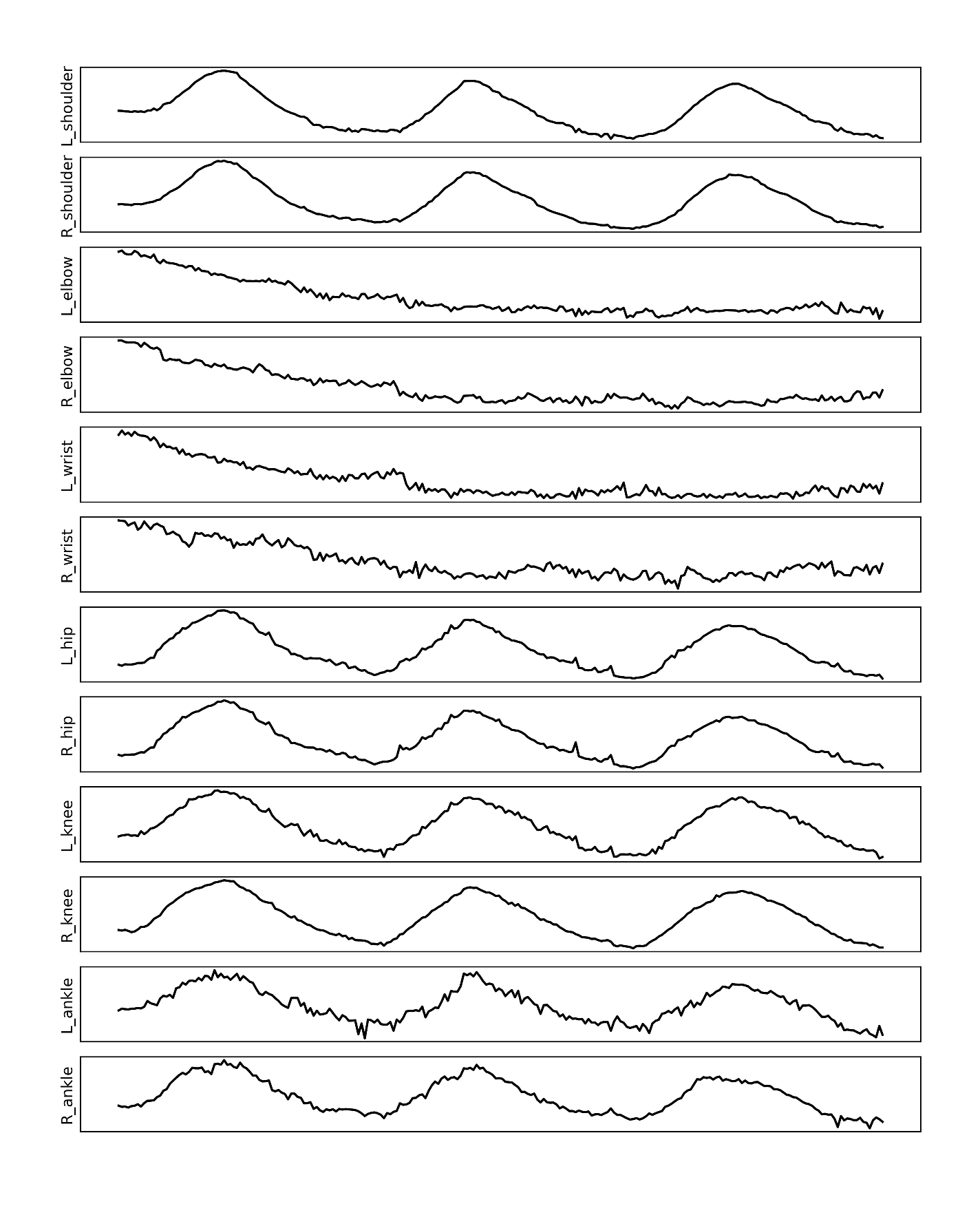} }}%
    \caption{Time series made of physical activity chin-ups for axis (a) x and (b) y}%
    \label{fig: time}%
\end{figure}

By running BlazePose in single-person pose estimation mode, we created data for each video frame containing 12 key points for the (x, y, z) axes. Since the data generation depends on each video frame, the generated data is time-series data in which the time is the number of frames per video (Figure \ref{fig: time}).

\subsection{Data Augmentation Using the Rotation Matrix}
\begin{figure}[ht!]
\includegraphics[width=\textwidth]{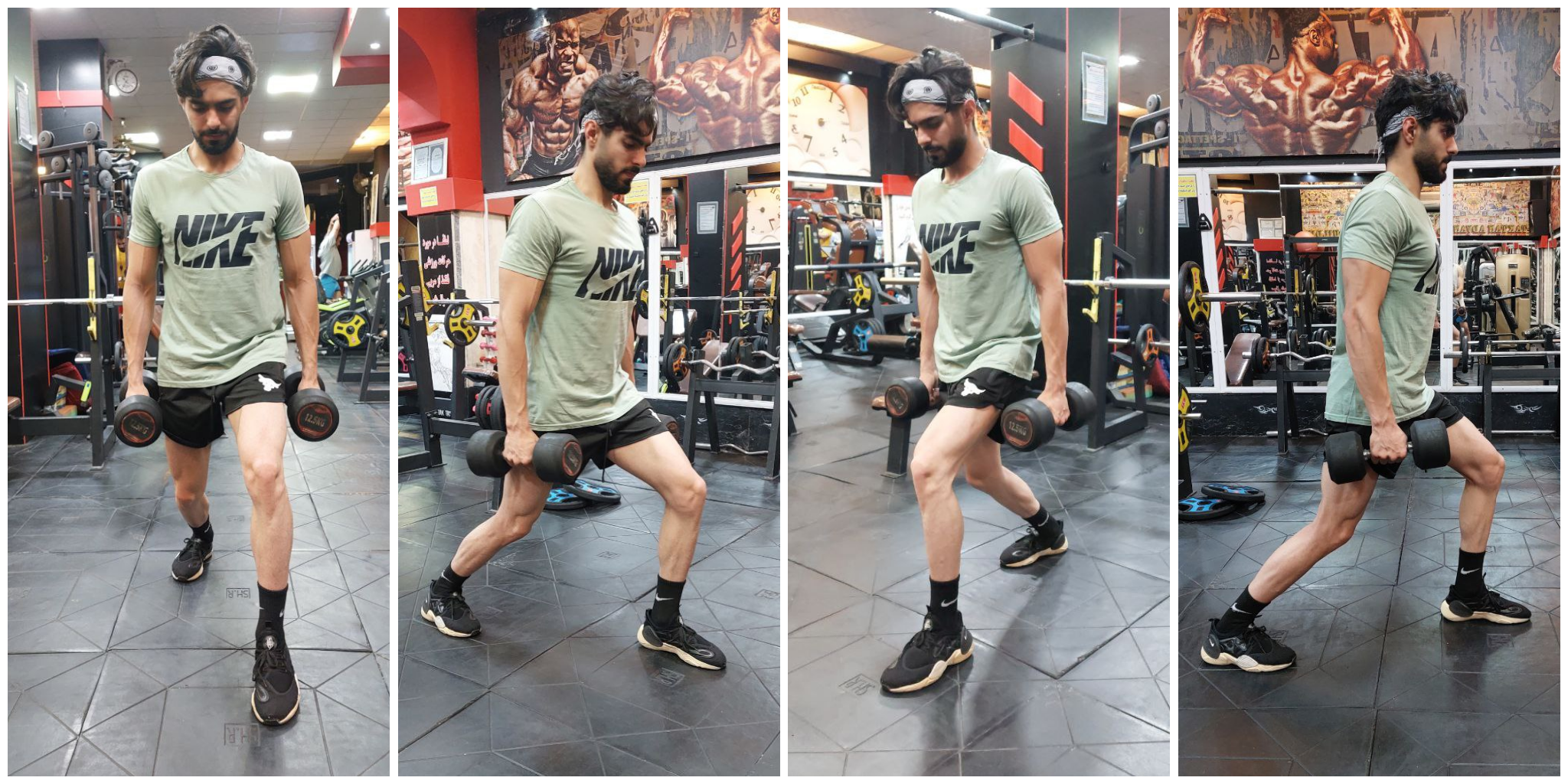}
      \centering
      \caption{Different images of dumbbell lunge from different camera perspectives}
      \label{fig: p}
\end{figure}

Since a person may be at different angles in front of the camera (See Figure \ref{fig: p}), this can cause ambiguity in detecting movements. In this study, we overcome the ambiguity in the camera perspective by using data augmentation based on \emph{Rotation Matrix}. By using the rotation matrix, different camera angles could be simulated, and this could help a classification algorithm to see different samples during its training time. 

\par
\begin{figure}[htbp]
\includegraphics[scale=0.9]{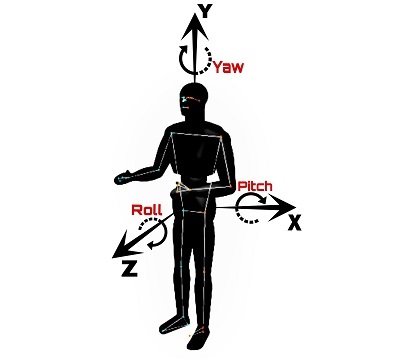}
      \centering
      \caption{Pitch, Roll, and Yaw describe the three degrees of freedom in the human body.}
      \label{fig: Pitch}
\end{figure}

Figure \ref{fig: Pitch} presents a shape of the human body that has three degrees of freedom: Pitch, Roll, and Yaw. However, in this study, we only use the degree of Yaw freedom for data augmentation. We rotate the image around the $y$-Axis by computing the following equation:
\begin{equation}
    R_y(\theta)=
\begin{bmatrix}
cos\theta & 0 & \sin\theta \\
0 & 1 & 0 \\
-\sin\theta & 0 & \cos\theta \\
\end{bmatrix}
\begin{bmatrix}
x \\
y \\
z \\
\end{bmatrix}
=
\begin{bmatrix}
x' \\
y' \\
z' \\
\end{bmatrix}
\end{equation}
In this equation $R_y(\theta)$ represents a rotation matrix around the y-axis. It is a 3x3 matrix that describes the transformation of coordinates after rotating them by an angle $\theta$ around the y-axis. $\theta$ represents the angle of rotation around the y-axis. $x$, $y$, and $z$ represent the original coordinates of a point in 3D space. $x'$, $y'$, and $z'$ represent the transformed coordinates after applying the rotation matrix $R_y(\theta)$ to the original coordinates $(x, y, z)$.
\par
\begin{figure}[htbp]
\includegraphics[width=\textwidth]{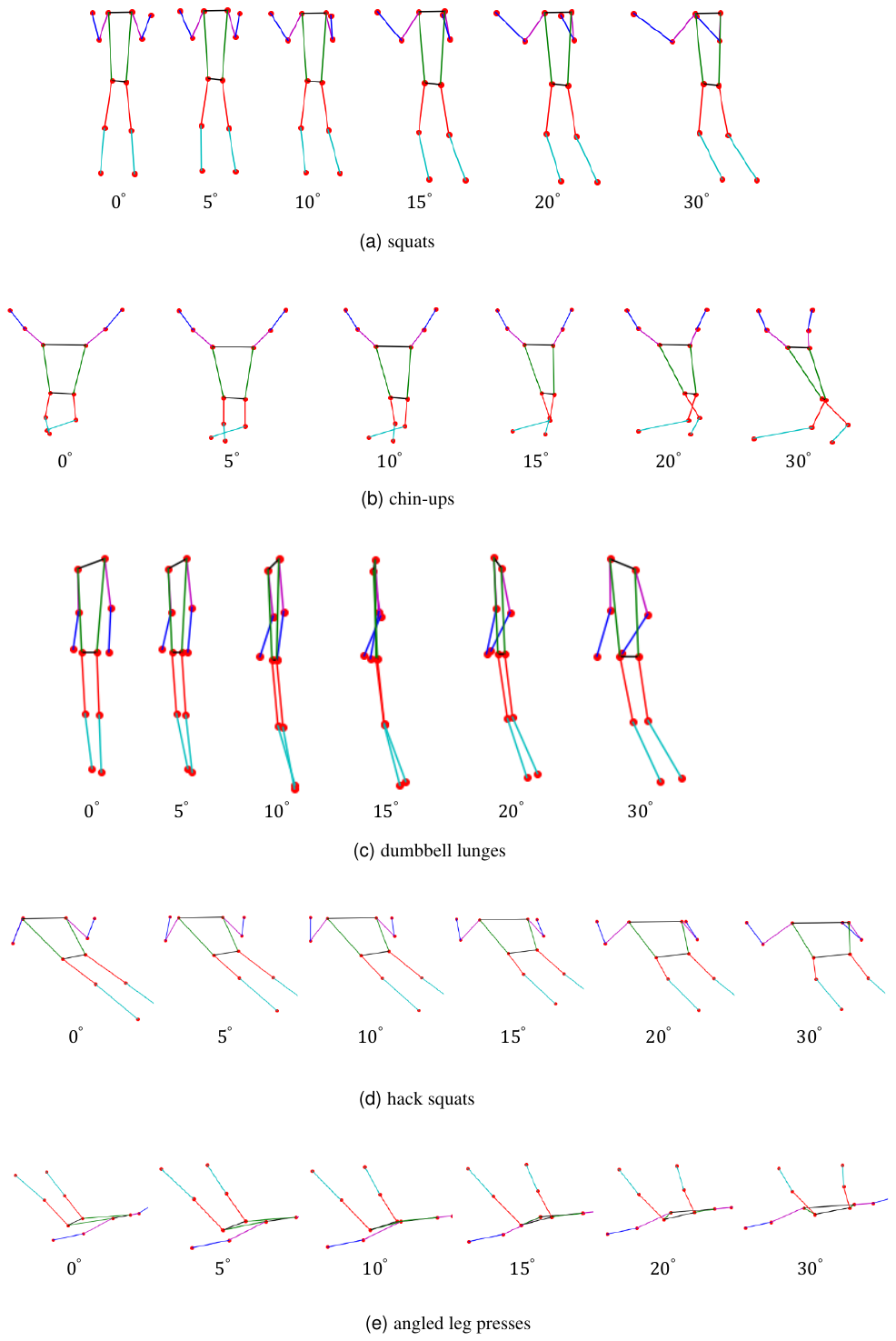}
      \centering
      \caption{Representation of the y-axis rotation at different angles of a video frame for physical activity (a) squats, (b) chin-ups, (c) dumbbell lunges, (d) hack squats, and (e) angled leg presses. The zero angle represents the pose in the original state, and the angles 5, 10, 15, 20, and 30 are the data generated from it.}
      \label{fig: Representation}
\end{figure}

To implement the data augmentation, we rotated $y$-Axis 180 degrees (increasing by one degree) clockwise and 180 degrees counterclockwise. Hence, in any view of the person in the proximity of the camera, we can have a different perspective in 360 degrees. This simple but effective data augmentation helps the algorithm identify and compare physical activities from different perspectives in front of the camera. Figure \ref{fig: Representation} is a representation of several different angles created for the 5 Physical Activities in the dataset. 
\par
Nevertheless, it should be noted that applying rotation could distort the body structure and real position of the body joints. Since many algorithm that benefits from pose estimation focus on identifying the movements and not exactly the position of joints, the distortion caused by rotation does not affect the accuracy of these algorithms. Therefore, we recommend tolerating this distortion for the sake of the data augmentation.  
Nevertheless, it should be noted that applying rotation could distort the body structure and real position of the body joints. Since many algorithms benefit from pose estimation and try to identify the movements and not exactly the position of joints, the distortion caused by rotation does not affect the accuracy of these algorithms. Therefore, we recommend tolerating this distortion for the sake of the data augmentation.  

\subsection{Data preprocessing and Feature Reduction}
In our study, the joints that are extracted from each video frame by the pose estimation algorithm are considered a feature. We have reduced the number of features for two main reasons, which are described as follows:

First, since each frame has 12 joints in three dimensions (x, y, z), the number of features to build a model has too many dimensions. To reduce the number of features in the first step, the z-dimensions were removed. 

A second reason is that because the person repeats a given exercise multiple times, each complete execution of the movement produces the same time series pattern. As a result, the type of movement can only be determined by having it repeated once on video. Hence, we limited the video frame count to 50. By limiting the number of frames, the number of redundant and repetitive features is reduced.

As a result, only the first 50 video frames of each of the extracted x and y dimensions were used. With 12 joints, each has x and y dimensions,  by selecting the first 50 frames of each sample, a time series with a length (number of features) of 1200 was constructed. 

However, the results of the experiments have shown that by removing the right shoulder feature, the classification algorithm improves the accuracy for detecting the type of exercise detection (The experimental results can be seen in section \ref{sec4}). Therefore, we used only 11 joints, and as a result, the number of features was reduced from 1200 to 1100.

\subsection{Evaluation Measures}
An important aspect that should be considered after creating a machine-learning model is how to generalize the model to unseen data. It must be ensured that the model is efficient and the results of its predictions can be trusted. This will be achieved via evaluation. To evaluate our approach, we have used the following performance metrics: accuracy, precision, recall, and F1-score. 

\section{EXPERIMENTAL RESULTS}
\subsection{Comparison Between the Models}
In the study, Logistic Regression, Gaussian Naive Bayes (GNB), Decision Tree, Support Vector Machine (SVM) with RBF kernel, K Nearest Neighbor (KNN), and Stochastic Gradient Descent on Support Vector Machine (SVM-SGD) were used for our experiments.
Classifiers were implemented using the scikit-learn package of Python \footnote{https://scikit-learn.org}.

\begin{figure}[htbp]
    \centering
    \subfloat[\centering No data Augmentation]{{\includegraphics[width=7cm]{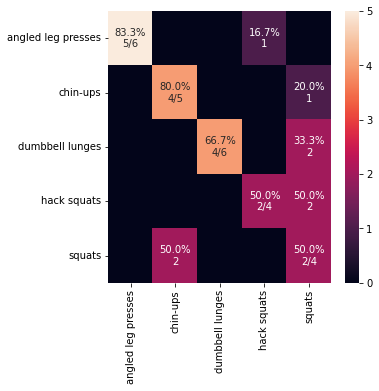} }}%
    \qquad
    \subfloat[\centering Pose data Augmentation]{{\includegraphics[width=7cm]{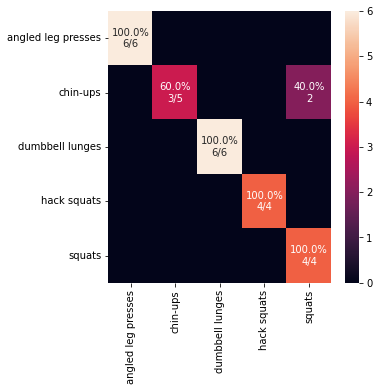} }}%
    \caption{Confusion matrix of Logistic Regression}%
    \label{fig: Logistic Regression}%
\end{figure}

\begin{figure}[htbp]
    \centering
    \subfloat[\centering No data Augmentation]{{\includegraphics[width=7cm]{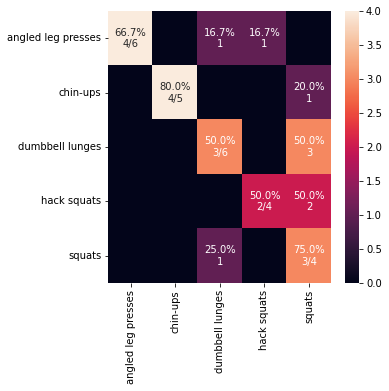} }}%
    \qquad
    \subfloat[\centering Pose data Augmentation]{{\includegraphics[width=7cm]{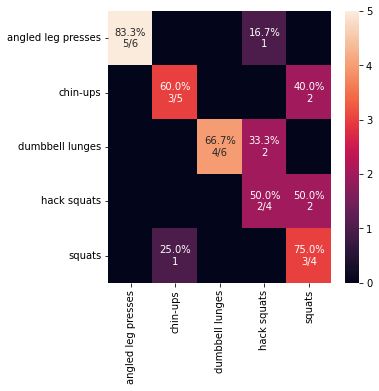} }}%
    \caption{Confusion matrix of Gaussian Naive Bayes }%
    \label{fig: Gaussian Naive Bayes }%
\end{figure}

\begin{figure}[htbp]
    \centering
    \subfloat[\centering No data Augmentation]{{\includegraphics[width=7cm]{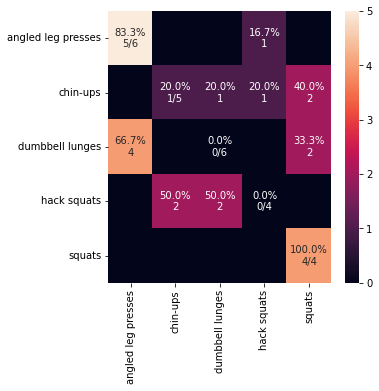} }}%
    \qquad
    \subfloat[\centering Pose data Augmentation]{{\includegraphics[width=7cm]{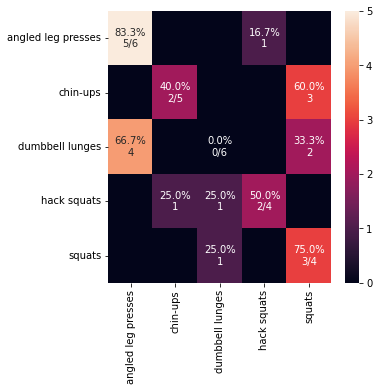} }}%
    \caption{Confusion matrix of DecisionTree }%
    \label{fig: DecisionTree }%
\end{figure}

\begin{figure}[htbp]
    \centering
    \subfloat[\centering No data Augmentation]{{\includegraphics[width=7cm]{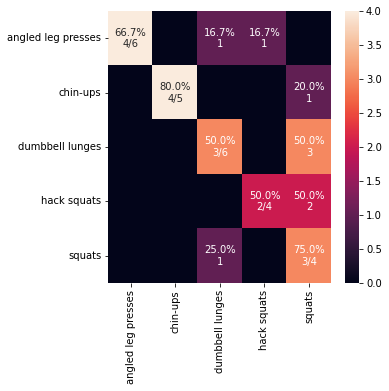} }}%
    \qquad
    \subfloat[\centering Pose data Augmentation]{{\includegraphics[width=7cm]{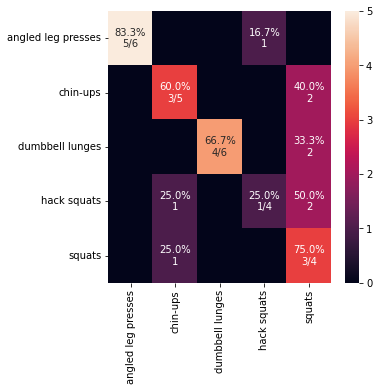} }}%
    \caption{Confusion matrix of SVM }%
    \label{fig: svm }%
\end{figure}

\begin{figure}[htbp]
    \centering
    \subfloat[\centering No data Augmentation]{{\includegraphics[width=7cm]{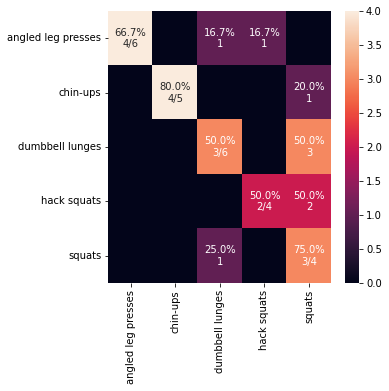} }}%
    \qquad
    \subfloat[\centering Pose data Augmentation]{{\includegraphics[width=7cm]{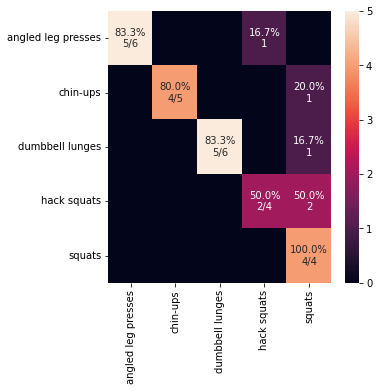} }}%
    \caption{Confusion matrix of KNN  }%
    \label{fig: KNN  }%
\end{figure}

\begin{figure}[htbp]
    \centering
    \subfloat[\centering No data Augmentation]{{\includegraphics[width=7cm]{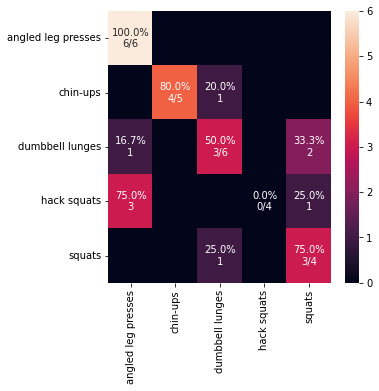} }}%
    \qquad
    \subfloat[\centering Pose data Augmentation]{{\includegraphics[width=7cm]{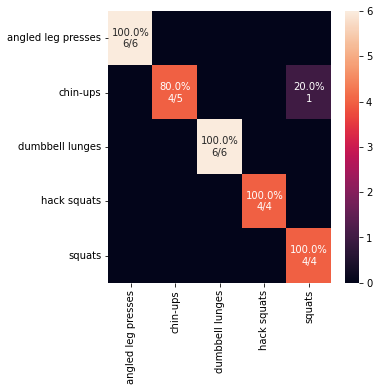} }}%
    \caption{Confusion matrix of SVM-SGD }%
    \label{fig: SVM-SGD}%
\end{figure}

Figures \ref{fig: Logistic Regression} to \ref{fig: SVM-SGD} show the confusion matrix of Logistic Regression, Gaussian Naive Bayes, Decision Tree, Support Vector Machines, K Nearest Neighbor (kNN), and SVM-SGD Classifier on no data augmentation and pose data augmentation. Table \ref{tab:Models} shows the results of our experiments for different models based on the accuracy, precision, recall, and F1-score metrics. As presented in Table \ref{tab:Models}, the SVM-SGD classifier has obtained the best result in all metrics using the Pose estimation method with a score of 96\%. Note the significant differences between augmented and not-augmented data in the quality of pose data classification. The SVM-SGD classifier, which is the top classifier, has 32\% accuracy using the pose data augmentation without augmentation versus using augmentation. As the confusion matrix in Figure \ref{fig: SVM-SGD} presents, this classifier has not been able to classify only one sample using the Pose data Augmentation method correctly; it has also predicted correctly in other test samples.
\begin{table}[htbp]
\centering
\caption{comparison of the models}
\label{tab:Models}
\begin{tabular}{|l|l|l|l|l|l|}
\hline
\textbf{Classifier}                              & \textbf{Method}                   & \textbf{Accuracy} & \textbf{F1-score} & \textbf{Recall} & \textbf{Precision} \\ \hline
\multirow{2}{*}{Logistic   Regression}  & No   data Augmentation   & 0.68     & 0.67     & 0.66   & 0.72      \\ \cline{2-6} 
                                        & Pose   data Augmentation & 0.92     & 0.91     & 0.92   & 0.93      \\ \hline
\multirow{2}{*}{Gaussian   Naive Bayes} & No   data Augmentation   & 0.64     & 0.65     & 0.64   & 0.72      \\ \cline{2-6} 
                                        & Pose   data Augmentation & 0.68     & 0.67     & 0.67   & 0.72      \\ \hline
\multirow{2}{*}{Decision   Tree}        & No   data Augmentation   & 0.40     & 0.32     & 0.41   & 0.28      \\ \cline{2-6} 
                                        & Pose   data Augmentation & 0.48     & 0.45     & 0.50   & 0.45      \\ \hline
\multirow{2}{*}{SVM}                    & No   data Augmentation   & 0.64     & 0.65     & 0.64   & 0.72      \\ \cline{2-6} 
                                        & Pose   data Augmentation & 0.64     & 0.62     & 0.62   & 0.69      \\ \hline
\multirow{2}{*}{KNN}                    & No   data Augmentation   & 0.64     & 0.65     & 0.64   & 0.72      \\ \cline{2-6} 
                                        & Pose   data Augmentation & 0.80     & 0.79     & 0.79   & 0.83      \\ \hline
\multirow{2}{*}{\textbf{SVM-SGD}}                & No   data Augmentation   & 0.64     & 0.56     & 0.61   & 0.54      \\ \cline{2-6} 
                                        & Pose   data Augmentation & \textbf{0.96}     & \textbf{0.96}     & \textbf{0.96}   & \textbf{0.96}      \\ \hline
\end{tabular}
\end{table}

\subsection{Key points impacts}\label{sec4}
Table \ref{tab:Models2} displays results obtained from listed algorithms, using different numbers of key points. It can be observed that the SVM\_SGD algorithm yields the best result with 11 key points, specifically when the right shoulder key point is eliminated from the total of 12 key points.  The removal of the right shoulder key point potentially improved the performance of the SVM\_SGD algorithm by reducing noise or enhancing the discriminatory power of the remaining key points. This strategic adjustment improved the algorithm's accuracy and effectiveness, as demonstrated in the results presented in Table \ref{tab:Models2}.
\begin{table}[htbp]
\centering
\caption{The results with different key points. "12" represents all key points, "11" represents the key points when the right shoulder is removed, "6 left" represents the key points on the left side of the body, and "6 right" represents the key points on the right side of the body.}
\label{tab:Models2}
\begin{tabular}{|l|l|llll|}
\hline
\multirow{3}{*}{Classifier}             & \multirow{3}{*}{Method}  & \multicolumn{4}{l|}{Accuracy}                                                                               \\ \cline{3-6} 
                                        &                          & \multicolumn{4}{l|}{The number of key points}                                                               \\ \cline{3-6} 
                                        &                          & \multicolumn{1}{l|}{\textbf{12}} & \multicolumn{1}{l|}{\textbf{11}} & \multicolumn{1}{l|}{\textbf{6 left}} & \textbf{6 right} \\ \hline
\multirow{2}{*}{Logistic   Regression}  & No   data Augmentation   & \multicolumn{1}{l|}{68} & \multicolumn{1}{l|}{68} & \multicolumn{1}{l|}{64}              & 60               \\ \cline{2-6} 
                                        & Pose   data Augmentation & \multicolumn{1}{l|}{88} & \multicolumn{1}{l|}{92} & \multicolumn{1}{l|}{84}              & 72               \\ \hline
\multirow{2}{*}{Gaussian   Naive Bayes} & No   data Augmentation   & \multicolumn{1}{l|}{68} & \multicolumn{1}{l|}{64} & \multicolumn{1}{l|}{60}              & 48               \\ \cline{2-6} 
                                        & Pose   data Augmentation & \multicolumn{1}{l|}{68} & \multicolumn{1}{l|}{68} & \multicolumn{1}{l|}{68}              & 64               \\ \hline
\multirow{2}{*}{Decision   Tree}        & No   data Augmentation   & \multicolumn{1}{l|}{48} & \multicolumn{1}{l|}{40} & \multicolumn{1}{l|}{52}              & 32               \\ \cline{2-6} 
                                        & Pose   data Augmentation & \multicolumn{1}{l|}{36} & \multicolumn{1}{l|}{48} & \multicolumn{1}{l|}{56}              & 40               \\ \hline
\multirow{2}{*}{SVM}                    & No   data Augmentation   & \multicolumn{1}{l|}{68} & \multicolumn{1}{l|}{64} & \multicolumn{1}{l|}{60}              & 48               \\ \cline{2-6} 
                                        & Pose   data Augmentation & \multicolumn{1}{l|}{64} & \multicolumn{1}{l|}{64} & \multicolumn{1}{l|}{72}              & 56               \\ \hline
\multirow{2}{*}{KNN}                    & No   data Augmentation   & \multicolumn{1}{l|}{68} & \multicolumn{1}{l|}{64} & \multicolumn{1}{l|}{60}              & 48               \\ \cline{2-6} 
                                        & Pose   data Augmentation & \multicolumn{1}{l|}{80} & \multicolumn{1}{l|}{80} & \multicolumn{1}{l|}{80}              & 64               \\ \hline
\multirow{2}{*}{SVM-SGD}                & No   data Augmentation   & \multicolumn{1}{l|}{56} & \multicolumn{1}{l|}{64} & \multicolumn{1}{l|}{28}              & 48               \\ \cline{2-6} 
                                        & Pose   data Augmentation & \multicolumn{1}{l|}{76} & \multicolumn{1}{l|}{\textbf{96}} & \multicolumn{1}{l|}{72}              & 64               \\ \hline
\end{tabular}
\end{table}
\subsection{Comparison with State-of-the-Art Models}
\cite{39} and \cite{40}, they have reached the accuracies of 99.91 and 99.38 respectively on the same data set. Although their dataset was trained and evaluated on a small number of samples, i.e. 80 videos, however, the movement recognition was focused on yoga movements and their recognition is easier compared to complex fitness movements.  \cite{43}, it reached an accuracy of 99 on the Fitness dataset. However, this accuracy was obtained only in 2 classes.  \cite{44}, a system is presented that analyzes the angle between pairs of limbs to detect errors and provide the user with the correct function. However, authors did not specified the accuracy of their system for 2D motion and their system has only been tested for 2 moves. 

\cite{11} and \cite{12}, the accuracies of 93.6 and 95.69 have been reached on the Fitness dataset, with the number of classes being 5 and 7, respectively. However, the number of video samples was very high. Our model has been able to reach an accuracy of 96 with far fewer videos in 5 classes. 

\section{CONCLUSIONS AND FUTURE WORK}
In this study, we propose a data augmentation for pose data extraction. We use pose estimation to extract body points (jpits) from each input image by using the BlazePose estimation algorithm. To overcome the ambiguity of the camera, we created 360 data for each instance of the training data using the Rotation Matrix. We created data 180 degrees clockwise and 180 degrees counterclockwise. Next, we developed and tested several different machine-learning models on a test dataset to evaluate the impact of data augmentation on the pose classification task. Experimental results showed that the Gradient Descent on the Support Vector Machines algorithm provides the best accuracy to classify data from human pose keypoints.

Although this model has a high ability to detect movements,  On the other hand, this model is limited to a small number of movements. To overcome the existing limitation, future research endeavors should concentrate on collecting more extensive and diverse datasets that encompass a broader spectrum of motions. This dataset expansion will play a crucial role in improving the model's performance by exposing it to various movement patterns. Consequently, the model will be able to learn and classify movements more effectively, leading to enhanced accuracy. In addition to dataset expansion, future work should also explore the combination of multiple data augmentation techniques. The dataset can be further enriched by incorporating various augmentation methods, such as rotation, scaling, flipping, and noise addition. This augmentation process introduces additional variations in the data, enabling the model to generalize better and improve its classification accuracy.

To achieve optimal results, future research should consider the synergistic combination of both data augmentation techniques and dataset expansion. By expanding the dataset and employing diverse augmentation strategies, the model's accuracy can be significantly improved, allowing for the classification of a wider range of movements.
%Bibliography
\bibliographystyle{abbrvnat}  
\bibliography{references}

\end{document}